\documentclass[10pt,twocolumn,letterpaper]{article}

\usepackage{cvpr}
\usepackage{times}
\usepackage{epsfig}
\usepackage{graphicx}
\usepackage{amsmath}
\usepackage{amssymb}
\usepackage{caption}
\usepackage{subcaption}
\usepackage{multirow}
\usepackage{paralist}
\usepackage{color}
\usepackage{xcolor}
\usepackage[export]{adjustbox}

\usepackage[pagebackref=true,breaklinks=true,letterpaper=true,colorlinks,bookmarks=false]{hyperref}

\cvprfinalcopy % *** Uncomment this line for the final submission

\def\cvprPaperID{10} % *** Enter the CVPR Paper ID here

\long\def\name#1{{\fontfamily{lmtt}\selectfont #1}} % font for name

\begin{document}

%%%%%%%%% TITLE
%\title{Transfering Deep Networks for Image Retrieval: Which layer should we use?}
\title{Exploiting Local Features from Deep Networks for Image Retrieval}

\author{Joe Yue-Hei Ng \quad 
    Fan Yang \quad
    Larry S. Davis \\
    University of Maryland, College Park\\
    {\tt\small \{yhng,fyang,lsd\}@umiacs.umd.edu}
}

\maketitle
%\ifcvprfinal\thispagestyle{empty}\fi
%%%%%%%%% ABSTRACT
\begin{abstract}
    Deep convolutional neural networks have been successfully applied to image classification tasks.
    When these same networks have been applied to image retrieval, the assumption has been made that the last layers would give the best performance, as they do in classification.
    We show that for instance-level image retrieval, lower layers often perform better than the last layers in convolutional neural networks.
    We present an approach for extracting convolutional features from different layers of the networks, and adopt VLAD encoding to encode features into a single vector for each image.
    We investigate the effect of different layers and scales of input images on the performance of convolutional features using the recent deep networks OxfordNet and GoogLeNet.
    Experiments demonstrate that intermediate layers or higher layers with finer scales produce better results for image retrieval, compared to the last layer.  
    When using compressed 128-D VLAD descriptors, our method obtains state-of-the-art results and outperforms other VLAD and CNN based approaches on two out of three test datasets.
    Our work provides guidance for transferring deep networks trained on image classification to image retrieval tasks.
\end{abstract}

\section{Introduction}
\label{sec:intro}

Image retrieval has been an active research topic for decades.
Most existing approaches adopt low-level visual features, \emph{i.e}., SIFT descriptors, and encode them using bag-of-words (BoW), vector locally aggregated descriptors (VLAD) or Fisher vectors (FV) and their variants.
Since SIFT descriptors capture local characteristics of objects, such as edges and corners, they are particularly suitable for matching local patterns of objects for instance-level image retrieval.

Recently, convolutional neural networks (CNNs) demonstrated excellent performance on image classification problems such as PASCAL VOC and ImageNet Large Scale Visual Recognition Challenge (ILSVRC)~\cite{krizhevsky2012imagenet,szegedy14going,Simonyan14c,DBLP:conf/eccv/ZeilerF14}.
By training multiple layers of convolutional filters, CNNs are capable to automatically learn complex features for object recognition and achieve superior performance compared to hand-crafted features.
A few works have suggested that CNNs trained for image classification tasks can be adopted to extract generic features for other visual recognition tasks~\cite{DBLP:conf/icml/DonahueJVHZTD14,razavian2014cnn,DBLP:conf/cvpr/OquabBLS14}.
Although several approaches have applied CNNs to extract generic features for image retrieval tasks and obtained promising results, a few questions still remain unaddressed.
First, by default CNNs are trained for classification tasks, where features from the final layer (or higher layers) are usually used for decision because they capture more semantic features for category-level classification.
However, local characteristics of objects at the instance level are not well preserved at higher levels.
Therefore, it is questionable whether it is best to directly extract features from the final layer or higher layers for instance-level image retrieval, where different objects from the same category need to be separated.
Second, most existing work assumes the size of a test image is the same as that of the training images.
However, different scales of input images may affect the behavior of convolutional layers as images pass through the network.
Only a few recent works attempt to investigate such effects on the performance of CNNs for image retrieval~\cite{gong2014multi,razavian2015visual}.

In view of the power of low-level features (\emph{i.e.}, SIFT) in preserving the local patterns of instances, and the success of CNN features in abstracting categorical information, we process CNN activations from lower to higher layers to construct a new feature for image retrieval by VLAD, although other encoding schemes can be readily applied.
Recent deep networks OxfordNet and GoogLeNet pre-trained on ImageNet database are used for evaluation.
We find that features from lower layers capture more local patterns of objects, and thus perform better than features from higher layers for instance-level image retrieval, which indicates that it is not the best choice to directly apply the final layer or higher layers that are designed for classification tasks to instance-level image retrieval.
In addition, we conduct further experiments by changing the scale of input images and using the same feature extraction and encoding methods.
It is surprising that the behavior of filters in each layer changes significantly with respect to the scale of input images. %
With input images of higher resolution, even the filters at higher layers effectively capture local characteristics of images as well, apart from semantic concepts of objects, thus producing better features and subsequent better retrieval results.

The contributions of this work are three-fold.
First, we design and conduct systematic and thorough experiments to investigate the performance of features from different layers and different scales of input test images in instance-level image retrieval.
%
%Second, we derive new convolutional features from CNNs by considering activations at each location of a feature map as a feature vector and encode them by VLAD.
Second, we introduce using VLAD encoding of local convolutional features from CNNs for image retrieval.
The new convolutional feature mimics the ability of SIFT descriptors to preserve local characteristics of objects, in addition to the well-known power of CNNs of capturing category-level information.
Our framework, based on the new features, outperforms other VLAD and CNN based approaches even with a relatively low-dimensional representation.
Finally, we provide insights as to why lower layers should be used for instance-level image retrieval rather than higher layers, while higher layers may achieve better performance for high resolution input images.

\section{Related Work}
\label{sec:related}

Traditional image retrieval approaches rely on hand-crafted features like SIFT descriptors, which are usually encoded into bag-of-words (BoW) histograms~\cite{DBLP:conf/iccv/SivicZ03}.
To increase the discriminative ability of SIFT descriptors, RootSIFT~\cite{DBLP:conf/cvpr/ArandjelovicZ12} was proposed to address the burstiness problem by using the Hellinger kernel on the original SIFT descriptors.
J{\'e}gou \etal~\cite{DBLP:journals/pami/JegouPDSPS12-vlad} proposed the vector locally aggregated descriptor (VLAD) to obtain a compact representation as a replacement for BoW histograms, which achieves good results while requiring less storage.
PCA and whitening~\cite{DBLP:conf/eccv/JegouC12}, signed square root (SSR) on VLAD vectors~\cite{DBLP:journals/pami/JegouPDSPS12-vlad} and intra-normalization~\cite{arandjelovic2013all} are later applied to the original VLAD descriptors to reduce noise and further boost performance.
Multi-VLAD~\cite{arandjelovic2013all} is based on constructing and matching VLAD features of multiple levels from an image to improve localization accuracy.
Other global features such as GIST descriptors and Fisher Vector (FV)~\cite{DBLP:conf/cvpr/PerronninLSP10} have also been evaluated for large-scale image retrieval.
Some approaches rely on semantic concepts or attributes to capture mid-level image information~\cite{DBLP:conf/cvpr/DouzeRS11,DBLP:conf/cvpr/SiddiquieFD11,DBLP:conf/cvpr/RastegariDPF13}, where attributes are binary values indicating the presence of semantic characteristics.
Relative attributes have been widely applied to refine search results.
In~\cite{DBLP:conf/cvpr/KovashkaPG12}, a set of ranking functions are learned offline to predict the strength of attributes, which are then updated by relative attribute feedback to rerank relevant images from the query stage.
Implicit feedback~\cite{DBLP:conf/iccv/ParikhG13} to learn ranking functions using implied user feedback cues and pivot attributes selection~\cite{DBLP:conf/iccv/KovashkaG13} to reduce the system's uncertainty have also been proposed to improve reranking performance.
~\cite{DBLP:conf/iccv/KovashkaG13a} learns a generic prediction function and adapts it into a user-specific function using user-labeled samples for personalized image search.

CNNs have led to major improvements in image classification~\cite{DBLP:conf/icml/DonahueJVHZTD14,razavian2014cnn,DBLP:conf/cvpr/OquabBLS14}.
As a universal image representation, CNN features can be applied to other recognition tasks and perform well~\cite{DBLP:conf/cvpr/OquabBLS14,DBLP:conf/icml/DonahueJVHZTD14,DBLP:conf/eccv/ZeilerF14}.
Razavian \etal~\cite{razavian2014cnn} first investigated the use of CNN features, \emph{i.e.}, OverFeat~\cite{DBLP:journals/corr/SermanetEZMFL13}, for various computer vision tasks, including image retrieval.
However, the performance of CNN feature extracted from the final layer lags behind that of simple SIFT-based methods with BoW and VLAD encoding.
Only by additionally incorporating spatial information do they achieve comparable results.
In~\cite{Crowley14a}, CNN features learned from natural images with various augmentation and pooling schemes are applied to painting retrieval and achieve good results.
Gong \etal~\cite{gong2014multi} introduce Multi-scale Orderless Pooling (MOP) to aggregate CNN activations from higher layers with VLAD, where these activations are extracted by a sliding window with multiple scales.
Experiments on an image retrieval dataset have shown promising results, but choosing which scales and layers to use remains unclear.
In~\cite{babenko2014neural}, a CNN model is retrained on a separate landmark database that is similar to the images at query time.
Not surprisingly, features extracted from the retrained CNN model obtain very good performance.
Unfortunately, collecting training samples and retraining the entire CNN model requires significant amounts of human and computing resources, making the application of this approach rather limited.
\cite{DBLP:conf/mm/WanWHWZZL14} conducted a comprehensive study on applying CNN features to real-world image retrieval with model retraining and similarity learning.
Encouraging experimental results show that CNN features are effective in bridging the semantic gap between low-level visual features and high-level concepts.
Recently, \cite{razavian2015visual} conducted extensive experiments on different instance retrieval dataset and obtained excellent results by using spatial search with CNN features.
Our work is inspired by \cite{gong2014multi} which also employs VLAD on CNN activations on multi-scale setting, but fundamentally different from \cite{gong2014multi}.
They utilize higher layers and multi-scale sliding window to extract CNN features from multiple patches independently, so the network has to be applied multiple times.
In contrast, we apply the network only once to the input image, and extract features at each location of the convolutional feature map in each layer.
We also explicitly verify the effectiveness of intermediate layers for image retrieval and provide additional analysis on the effect of scale.

\cite{DBLP:journals/corr/XuYH14} introduces latent concept descriptors for video event detection by extracting and encoding features using VLAD at the last convolutional layer with spatial pooling.
In contrast, we extend the use of convolutional features to lower layers without additional pooling to preserve local information.
We also focus on evaluating performance of different convolutional layers for instance-level image retrieval.

%CNN is excellent in image classification tasks, and various previous work tested it on image retrieval tasks.
%\cite{razavian2014cnn} founds CNN gives excellent performance using Overfeat.
%Our work is similar to~\cite{gong2014multi}, which also employs VLAD to do multi-scale orderless pooling on CNN features.
%Their work focus on using the last fully connected layer.
%This work provide analysis and discussion on their method on various scale and layers.
%
%This paper is also related to a line of work that study the transferability of network.
%CVPR14 paper shows that mid level representation can be transferred to various dataset and performs very well.
%NIPS14 paper shows a study of the effect of fine-tuning on different layers of network.
%Our work focus on image retrieval tasks and use a pre-trained network without fine-tuning.

\section{Approach}
\label{sec:approach}

We describe our approach of extracting and encoding CNN features for image retrieval in this section.
We start by introducing the deep neural networks used in our framework, and then describe the method for extracting features.
To encode features for efficient retrieval, we adopt VLAD to compress the CNN features into a compact representations.

\subsection{Convolutional neural network}
\label{sec:network}

Our approach is applicable to various convolutional neural network architectures.
We experiment with two variants of recent deep neural networks: OxfordNet~\cite{Simonyan14c} and GoogLeNet~\cite{szegedy14going}, which ranked top two in ILSVRC 2014.
The networks are pre-trained on ImageNet by Caffe implementation~\cite{jia2014caffe} and publicly available on the Caffe model zoo.
We adopt the 16 layers OxfordNet trained by~\cite{Simonyan14c} as it gives similar performance to the 19 layer version.
The network consists of stacked $3 \times 3$ convolutional layers and pooling layers, followed by two fully connected layers and takes images of $224 \times 224$ pixels as input.
We also use a 22-layer deep convolutional network GoogLeNet~\cite{szegedy14going}, which gives state-of-the-art results in ImageNet classification tasks.
The GoogLeNet takes images of $224 \times 224$ pixels as input that is then passed through multiple convolutional layers and stacking ``inception'' modules.
Each inception module is regarded as a convolutional layer containing $1 \times 1$, $3 \times 3$ and $5 \times 5$ convolutions, which are concatenated with an additional $3 \times 3$ max pooling, with $1 \times 1$ convolutional layers in between for dimensionality reduction.
There are totally 9 inception modules sequentially connected, followed by an average pooling and a softmax at the end.
Unlike OxfordNet, fully connected layers are eliminated which simplifies our experiments, so that we can focus on the convolutional feature maps.
Finally, the networks are trained by average-pooled activation followed by softmax.
The fully convolutional network GoogLeNet simplifies the extension to applying the network to multiple scales of images, and lets us encode the local convolutional features in the same way for all layers, which allows fair comparisons among layers.
Table~\ref{table:networks} shows the output size of intermediate layers in OxfordNet and GoogLeNet.
Since it is time consuming to evaluate the lower layers which have large feature maps, some lower layers are omitted in our evaluation.

\subsection{Extracting convolutional features}

Given a pre-trained network (OxfordNet or GoogLeNet) with $L$ layers, an input image $\mathcal{I}$ is first warped into an $n \times n$ square to fit the size of training images, and then is passed through the network in a forward pass.
In the $l$-th convolutional layer $\mathcal{L}_l$, after applying the filters to the input image $\mathcal{I}$, we obtain an $n^l \times n^l \times d^l$ feature map $\mathcal{M}^l$, where $d^l$ is the number of filters with respect to $\mathcal{L}_l$.
For notational simplicity, we denote $n_s^l = n^l \times n^l$.
Similar to the strategy in \cite{DBLP:journals/corr/XuYH14}, at each location $(i,j), 1 \le i \le n^l$ and $1 \le j \le n^l$, in the feature map $\mathcal{M}^l$, we obtain a $d^l$-dimensional vector $\mathbf{f}_{i,j}^l \in \mathbb{R}^{d^l}$ containing activations of all filters, which is considered as our feature vector.
In this way, we obtain $n_s^l$ local feature vectors for each input image at the convolutional layer $\mathcal{L}_l$, denoted as $\mathbf{F}^l = \{ \mathbf{f}_{1,1}^l, \mathbf{f}_{1,2}^l, \cdot \cdot \cdot, \mathbf{f}_{n^l,n^l}^l \} \in \mathbb{R}^{d^l \times n_s^l}$.
While \cite{DBLP:journals/corr/XuYH14} only extracts features from the last convolutional layer, we extend the feature extraction approach to all convolutional layers.
By processing the input image $\mathcal{I}$ throughout the network, we finally obtain a set of feature vectors for each layer, $\{ \mathbf{F}^1, \mathbf{F}^2, \cdot \cdot \cdot, \mathbf{F}^L \}$.
The feature extraction procedure is illustrated in Figure~\ref{fig:overview}\footnote{The k-means clustering figure is from \url{http://www.vlfeat.org/overview/kmeans.html}}.

\subsection{VLAD encoding}
Unlike image classification, which is trained with many labeled data for every category, in instance retrieval generally there is no training data available.
Therefore, a pre-trained network is likely to fail to produce good holistic representations that are invariant to translation or viewpoint changes while preserving instance level information.
In contrast, local features, which focus on smaller parts of images, are easier to represent and generalize to other object categories while capturing invariance.

%\textcolor{gray}{\small
%The convolutional feature map at lower layers has larger size, thus having higher unrolled dimension than the feature map at higher layers.
%
%For example, the feature from \name{Inception 3a} layer of GoogLeNet has $28 \times 28 \times 256 = 200704$ dimensions.
%%
%In contrast, the feature from \name{Inception 5a} layer has $7 \times 7 \times 832 = 40768$ dimensions.
%
%With such high dimensionality, finding nearest neighbors of an image by directly computing the L2 distance between sets of feature vectors is inefficient.
%
%Although average pooling and max pooling are used in convolutional networks to reduce the dimensionality of convolutional features from intermediate layers, they focus more on capturing category-level abstraction rather than preserving local details of particular objects.
%
%Therefore, directly adopting the lower-dimensional features from higher layers is not a good choice.
%}

\begin{figure}[ht]
\begin{center}
    \includegraphics[width=\linewidth]{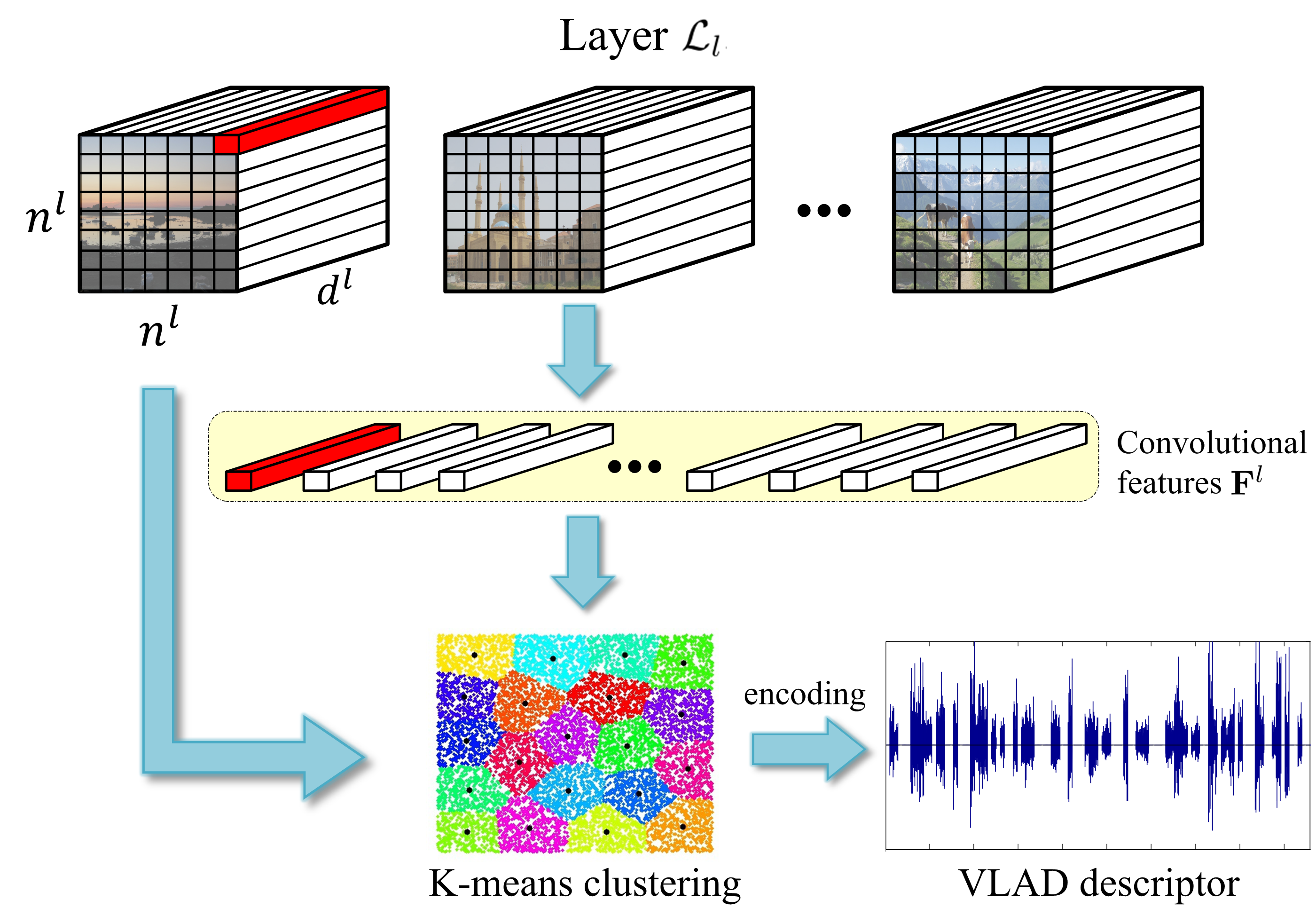}
\caption{Overview of our feature extraction and encoding.}
    \label{fig:overview}
\end{center}
\end{figure}

Since each image contains a set of low-dimensional feature vectors, which has similar structure as dense SIFT, we propose to encode these feature vectors into a single feature vector using standard VLAD encoding.
The VLAD encoding is effective for encoding local features into a single descriptor while achieving a favorable trade-off between retrieval accuracy and memory footprint.
An overview of our system is illustrated in Figure~\ref{fig:overview}.

\begin{table}[th]
    \vspace{-1.0em}
{\small
\centering
\begin{subtable}[b]{0.45\textwidth}
\centering
\begin{tabular}{ c c }
\hline
Layer (low $\rightarrow$ high) & Output size ($n^l \times n^l \times d^l$) \\
\hline\hline
\name{pool1-norm1} & $56\times56\times64$ \\
\name{conv2-norm2} & $28\times28\times192$ \\
\hline
\name{Inception 3a} & $28\times28\times256$ \\
\name{Inception 3b} & $28\times28\times480$ \\
\hline
\name{Inception 4a} & $14\times14\times512$ \\
\name{Inception 4b} & $14\times14\times512$ \\
\name{Inception 4c} & $14\times14\times512$ \\
\name{Inception 4d} & $14\times14\times528$ \\
\name{Inception 4e} & $14\times14\times832$ \\
\hline
\name{Inception 5a} & $7\times7\times832$ \\
\name{Inception 5b} & $7\times7\times1024$ \\
\hline
\end{tabular}
\caption{GoogLeNet}
\vspace{10px}
\begin{tabular}{ c c }
\hline
Layer (low $\rightarrow$ high) & Output size ($n^l \times n^l \times d^l$) \\
\hline\hline
\name{conv2\_1} & $112\times112\times128$ \\
\name{conv2\_2} & $112\times112\times128$ \\
\name{conv2\_3} & $112\times112\times128$ \\
\hline
\name{conv3\_1} & $56\times56\times256$ \\
\name{conv3\_2} & $56\times56\times256$ \\
\name{conv3\_3} & $56\times56\times256$ \\
\hline
\name{conv4\_1} & $28\times28\times512$ \\
\name{conv4\_2} & $28\times28\times512$ \\
\name{conv4\_3} & $28\times28\times512$ \\
\hline
\name{conv5\_1} & $14\times14\times512$ \\
\name{conv5\_2} & $14\times14\times512$ \\
\name{conv5\_3} & $14\times14\times512$ \\
\hline
\end{tabular}
\caption{OxfordNet}
\end{subtable}
}
\caption{Size of feature maps}
\label{table:networks}
\end{table}

VLAD encoding is similar to constructing BoW histograms.
Given a collection of L2-normalized convolutional features from layer $\mathcal{L}_l$, we perform k-means clustering to obtain a vocabulary $\mathbf{c}_1^l, ..., \mathbf{c}_k^l$ of $k$ visual words, where $k$ is relatively small ($k = 100$ in our experiments following~\cite{gong2014multi}), so the vocabulary is coarse.
For each image, a convolutional feature $\mathbf{f}_{i,j}^l$ from layer $\mathcal{L}_l$ is assigned to its nearest visual word $\mathbf{c}_i^l = NN(\mathbf{f}_{i,j}^l)$.
For the visual word $\mathbf{c}_i^l$, the vector difference between $\mathbf{c}_i^l$ and the feature $\mathbf{f}_{i,j}^l$ (residual), $\mathbf{f}_{i,j}^l - \mathbf{c}_i^l$, is recorded and accumulated for all features assigned to $\mathbf{c}_i^l$.
The VLAD encoding converts the set of convolutional features of an image, $\mathbf{F}^l$, from layer $\mathcal{L}_l$ to a single $d^l \times k$-dimensional vector $\mathbf{v}^l \in \mathbb{R}^{d^l \times k}$, describing the distribution of feature vectors regarding the visual words.
Formally, a VLAD descriptor of an image regarding layer $\mathcal{L}_l$ is represented as
\begin{equation}
\mathbf {v}^l = [ \sum_{NN(\mathbf{f}_{i,j}^l)=\mathbf{c}_1^l} \mathbf{f}_{i,j}^l-\mathbf{c}_1^l, \cdot \cdot \cdot, \sum_{NN(\mathbf{f}_{i,j}^l)=\mathbf{c}_k^l} \mathbf{f}_{i,j}^l-\mathbf{c}_k^l].
\end{equation}
Here $\sum_{NN(\mathbf{f}_{i,j}^l)=\mathbf{c}_k^l} \mathbf{f}_{i,j}^l-\mathbf{c}_k^l$ is the accumulated residual between the visual word $\mathbf{c}_k^l$ and all convolutional features $\mathbf{f}_{i,j}^l$ that are assigned to $\mathbf{c}_k^l$.
The VLAD descriptors are normalized by intra-normalization which has been shown to give superior results than signed square root (SSR) normalization~\cite{arandjelovic2013all}.
Since the dimensionality of the original VLAD descriptor is very high, making direct comparison expensive, we further apply PCA to reduce the dimensionality of VLAD descriptors to improve retrieval efficiency and then whitening to increase its robustness against noise.

\subsection{Image Retrieval}
For all database images and a query image, we extract convolutional features and encode them into VLAD descriptors.
Image retrieval is done by calculating the L2 distance between the VLAD descriptors of the query image and database images.
We use PCA to compress the original VLAD descriptors to relatively low-dimensional vectors (128-D), so that the computation of L2 distance can be done efficiently.
We will show in the experiments that the compressed 128-D VLAD vectors achieve excellent results with little loss of performance.
%

%Each layer is processed and compared independently.
%%
%Nevertheless, it is unclear which layer performs the best on different datasets due to large variance of database images.
%%
%To further improve the performance, we attempt to fuse the VLAD descriptors from different layers to exploit the complementary information.
%%
%Specifically, we combine multiple ranked lists from different layers to obtain final results, rather than directly concatenating VLAD descriptors.

\section{Experiments}
\label{sec:experiments}

We perform experiments on 3 instance-level image retrieval datasets: Holidays~\cite{Jegou2008hamming}, Oxford~\cite{DBLP:conf/cvpr/PhilbinCISZ07} and Paris~\cite{DBLP:conf/cvpr/PhilbinCISZ08}.
The Holidays dataset includes 1491 images of personal holiday photos from 500 categories, where the first image in each category is used as the query.
%
%Given a query, all the other 1490 images are treated as the database images.
%
The Oxford and Paris datasets consist of 5062 images and 6412 images of famous landmarks in Oxford and Paris, respectively.
%
%Both datasets have 55 queries from different landmarks (11 for Oxford and 12 for Paris) with specified rectangular region of interest enclosing the instance to be retrieved, where each landmark has multiple query images.
Both datasets have 55 queries with specified rectangular region of interest enclosing the instance to be retrieved, where each landmark has multiple query images.
To simplify the experiments, the rectangular regions are ignored and full images are used for retrieval in this work.
Following the standard evaluation protocol, we use mean average precision (mAP) to evaluate the performance of our approach.

\subsection{Comparison of layers}
\begin{figure*}[ht]
\begin{center}
    \begin{subfigure}[t]{0.42\linewidth}
        \includegraphics[width=0.95\linewidth, valign=t]{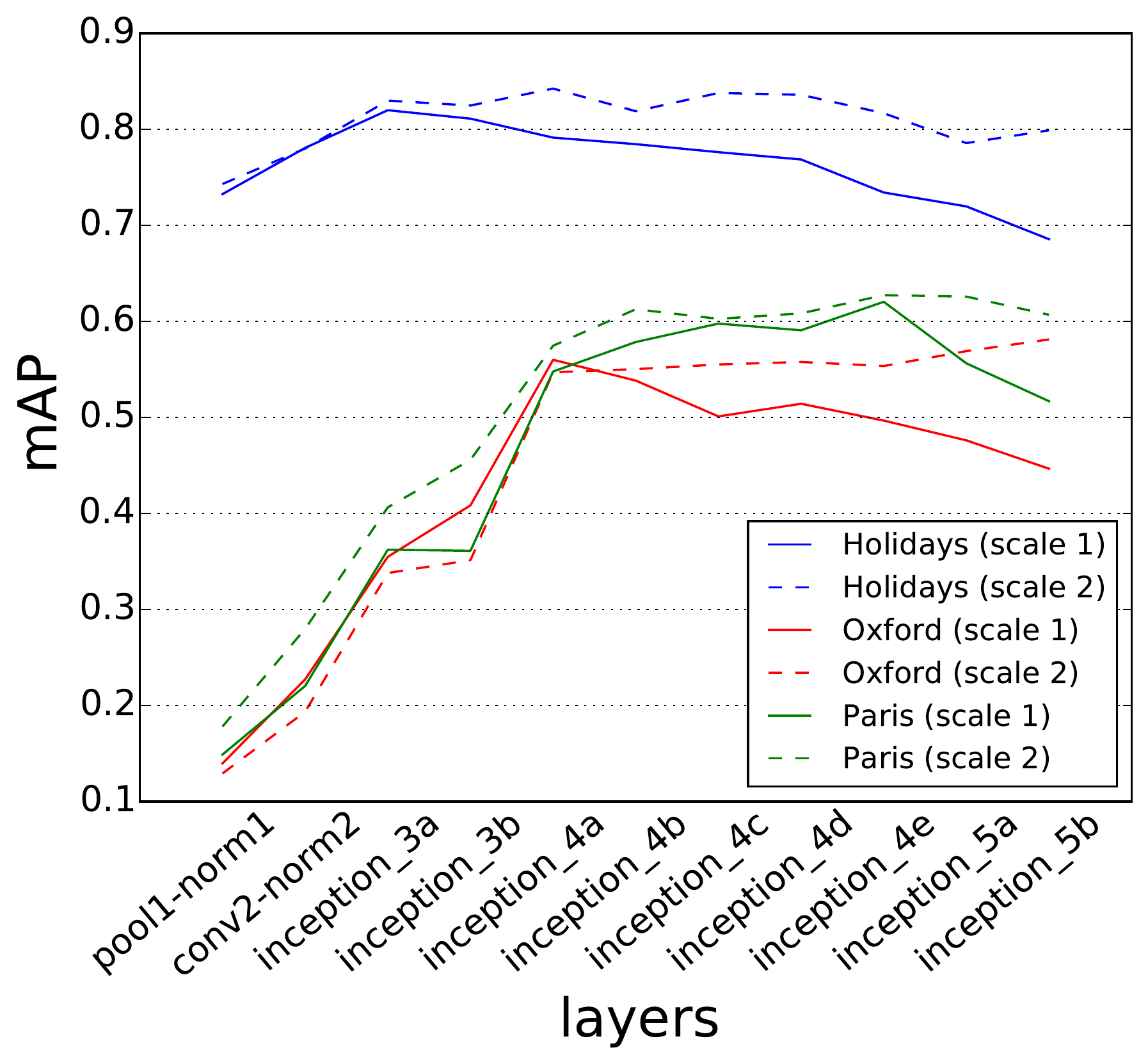}
        \caption{GoogLeNet}
    \end{subfigure}
    \begin{subfigure}[t]{0.42\linewidth}
        \includegraphics[width=0.95\linewidth, valign=t]{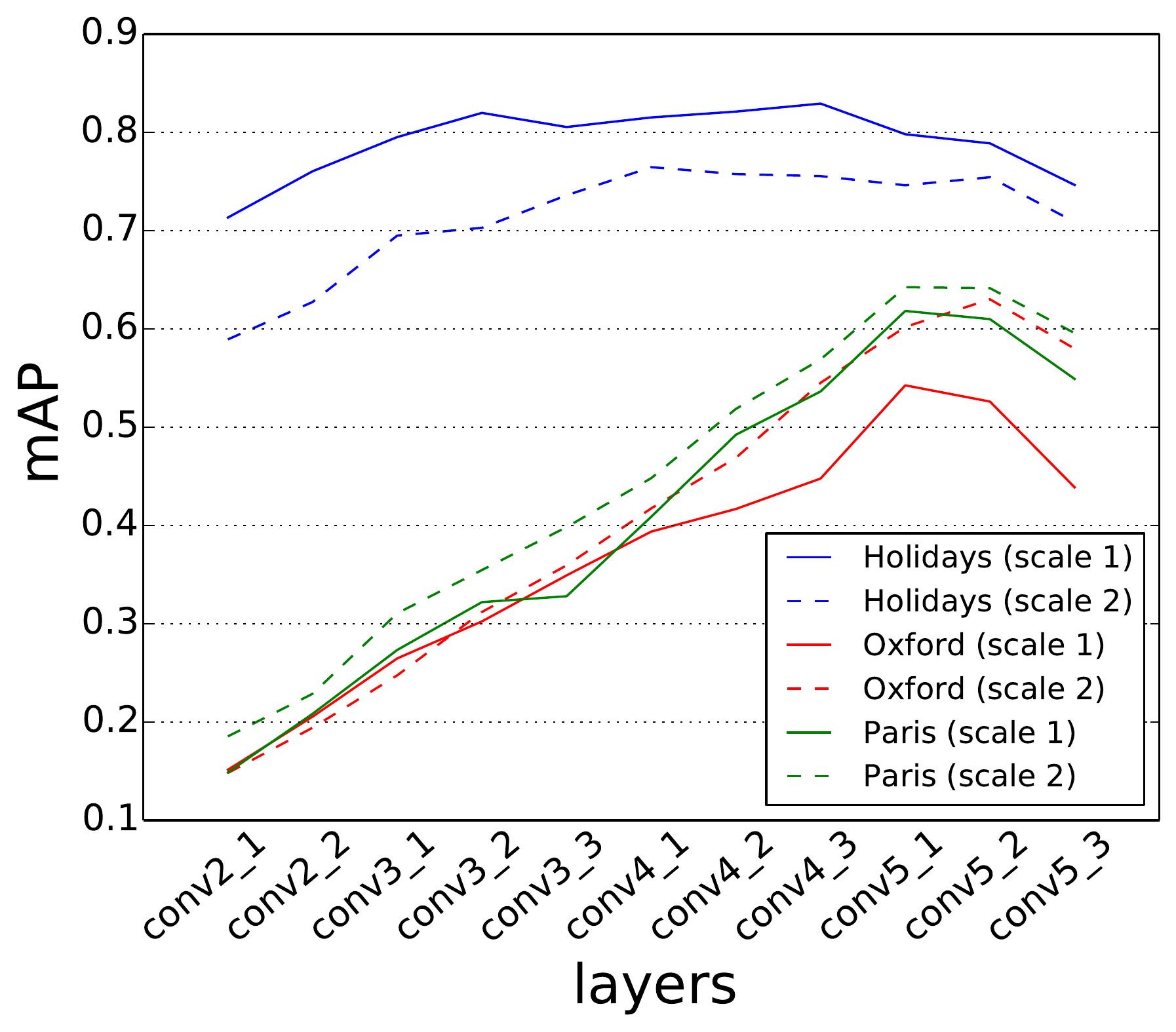}
        \caption{OxfordNet}
    \end{subfigure}

    \caption{Performance of different layers on both scales: Solid and dash lines correspond to the original and second scale respectively.
             Fully-connected layers of OxfordNet are omitted due to incompatible size of the last convolutional layer at scale 2.}
    \label{fig:layers}
\end{center}
\end{figure*}

We first study the performance of convolutional features from different layers.
We use VLAD to encode convolutional features from each layer and evaluate the mAP with respect to the corresponding layer.
Figure~\ref{fig:layers} shows the performance for both OxfordNet and GoogLeNet.
There is a clear trend in the results of both networks on the first scale (solid lines in the figure).
The mAP first increases as we go deeper into the network because the convolutional features achieve more invariance, until reaching a peak.
However, the performance at higher layers gradually drops since the features are becoming too generalized and less discriminative for instance-level retrieval.
The best performing layers of GoogLeNet on the Holidays, Oxford and Paris datasets are \name{Inception 3a}, \name{Inception 4a}, and \name{Inception 4e} respectively.
On the Holidays dataset, the performance of intermediate layers is much better than that of the last layer (82.0\% vs 68.5\%).
In contrast, the best performing layers on the Oxford and Paris datasets are from middle upper layers.
%
%\commentj{the name of the middle upper layers}
%
Nevertheless, similar trends can still be clearly seen on these two datasets that the intermediate layers perform better than the last layer.
We then conduct similar experiment with the 16 layers OxfordNet.
Although OxfordNet is less deeper than GoogLeNet, we still see this trend. %, especially when going up from convolutional layer \name{conv5\_1} to higher layers.
On the Oxford and Paris datasets, the best performing layer is not the last layer, but the intermediate convolutional layers \name{conv5\_1}, showing that increasing generalization at higher layers is not always useful in instance retrieval.
This verifies that across different network architectures and datasets, intermediate layers perform the best and should be used for instance-level retrieval.

When convolutional networks grow deeper, which gives an increasing number of choice for layers to transfer, it becomes more important to examine the layers used for image retrieval, since the layers perform very differently in deep networks.
Unlike recent work, which suggests only using the last two fully connected layers~\cite{razavian2014cnn,gong2014multi,babenko2014neural}, or the last convolutional layers~\cite{razavian2015visual}, our experiments show that higher layers are not always optimal depending on the tasks considered, especially for the very deep networks recently proposed.
For instance-level image retrieval, which is very different from classification tasks, lower layers usually perform better than higher layers as features from lower layers preserve more local and instance-level characteristics of objects.
We envisage this trend will become more pronounced when networks become deeper in the future.

\subsection{Scales}

Applying a network at multiple scales gives significant improvement over its original scale as shown in previous work~\cite{gong2014multi, razavian2014cnn}.
%We pass through $448 \times 448$ images into GoogLeNet, giving 4 times larger feature map on each layer, and examine the performance changes relative to the first scale.
In view of this, apart from using the original size of input images (scale 1), we enlarge the size of the input image to $2n \times 2n$ (scale 2) to generate 4 times larger feature maps at each layer, and conduct similar experiments.
We evaluate the difference in performance using features extracted from scale 1 and scale 2.

Figure~\ref{fig:layers} shows the performance of different layers at both scales.
In general, features from the finer scale, which are obtained from higher resolution images, give better performance than the original scale except OxfordNet on the Holidays dataset.
Interestingly, the relative performance among layers at the higher scale are quite different from the original scale from GoogLeNet.
On the Holidays dataset, the performance at scale 2 first increases and then decreases as we go up to higher layers.
The trend is similar to scale 1 although the performance difference between layers at scale 2 is smaller.
On the Oxford and Paris datasets, we obtain better results using features from higher layers than those from lower layers on the finer scale (scale 2).
%
%This confirms the similar experiment in a recent paper~\cite{razavian2015visual} that using higher resolution images is better in Oxford dataset but not in Holidays dataset, as Oxford dataset focuses on smaller objects.
%The different behaviors between datasets may be Oxford dataset is harder dataset and requires more invariance on the feature in matching process.
%
It is surprising that the networks perform better with larger input images, although by default they should take images of $224\times224$ pixels that they are trained on as the input~\cite{razavian2015visual}.
An intuitive explanation for the good performance of the last layer at scale 2 is that the original filters focus more on local details of enlarged images since the size of the filters remains unchanged.
Therefore, the convolutional features extracted from the higher layers at a finer scale actually focuses on smaller parts of the images, thus preserving mid-level details of objects to some extent instead of global categorical and abstract information as in the original scale.
Our experiments suggest that higher resolution images are preferable even if the network was trained at a coarser level.
In contrast, different layers in OxfordNet, which was trained in a multi-scale setting, behave similarly for both scales.
\begin{figure*}[ht]
\begin{center}
    \begin{tabular}{c c c c c}
        Original images & \name{Inception 4a} & \name{Inception 5b} & \name{Inception 5b} (scale 2) & \\
%        \includegraphics[width=0.2\linewidth]{figures/reconstruction/100000.png} &
%        \includegraphics[width=0.2\linewidth]{figures/reconstruction/100000-inception_4a-output-s1-nn1.png} &
%        \includegraphics[width=0.2\linewidth]{figures/reconstruction/100000-inception_5b-output-s1-nn1.png} &
%        \includegraphics[width=0.2\linewidth]{figures/reconstruction/100000-inception_5b-output-s2-nn1.png} &
%        1-NN \\
%         &
%        \includegraphics[width=0.2\linewidth]{figures/reconstruction/100000-inception_4a-output-s1-nn5.png} &
%        \includegraphics[width=0.2\linewidth]{figures/reconstruction/100000-inception_5b-output-s1-nn5.png} &
%        \includegraphics[width=0.2\linewidth]{figures/reconstruction/100000-inception_5b-output-s2-nn5.png} &
%        5-NN \\
%
        \includegraphics[width=0.155\linewidth]{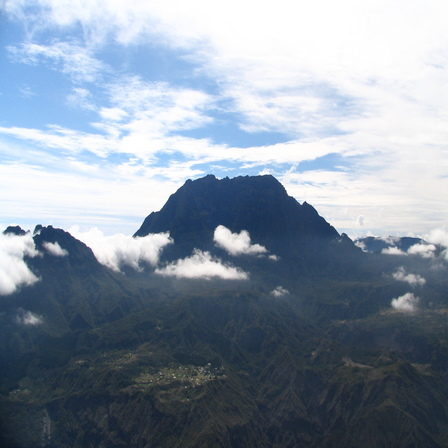} &
        \includegraphics[width=0.155\linewidth]{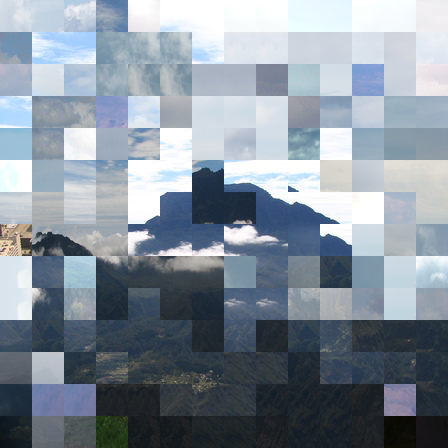} &
        \includegraphics[width=0.155\linewidth]{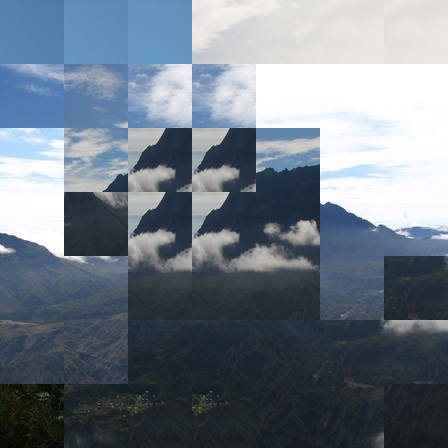} &
        \includegraphics[width=0.155\linewidth]{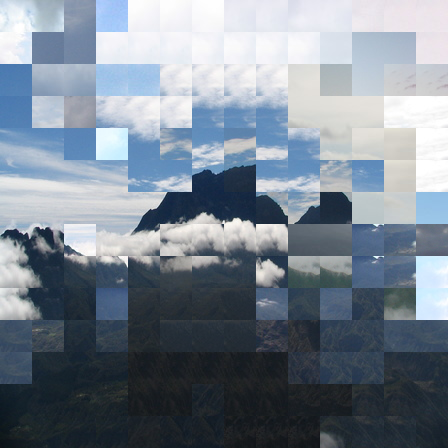} &
        \small{1-NN} \\
         &
        \includegraphics[width=0.155\linewidth]{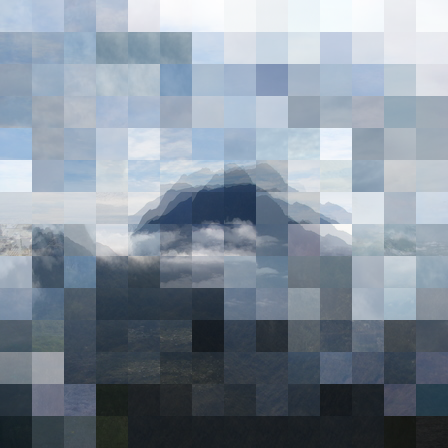} &
        \includegraphics[width=0.155\linewidth]{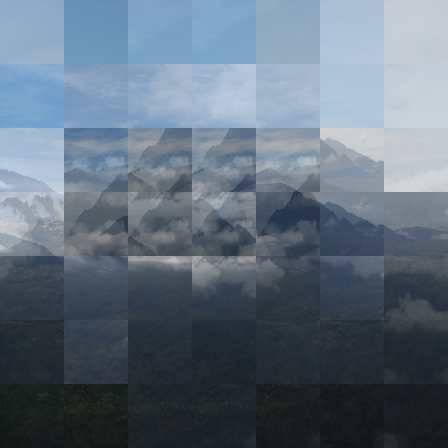} &
        \includegraphics[width=0.155\linewidth]{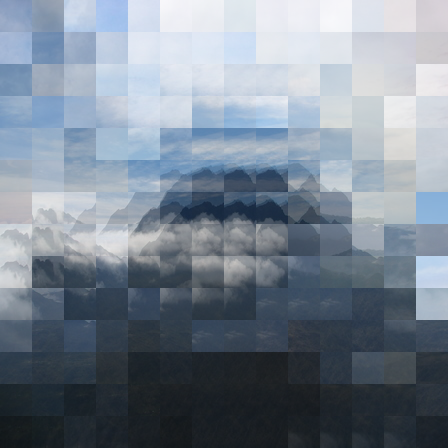} &
        \small{5-NN} \\

        \includegraphics[width=0.155\linewidth]{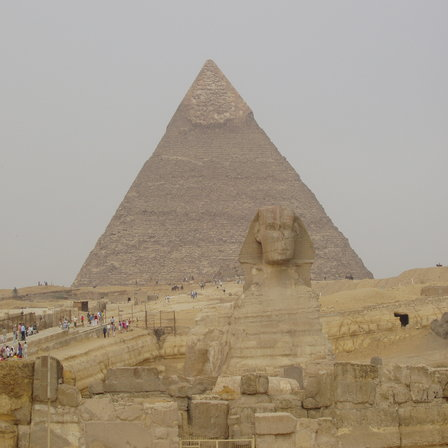} &
        \includegraphics[width=0.155\linewidth]{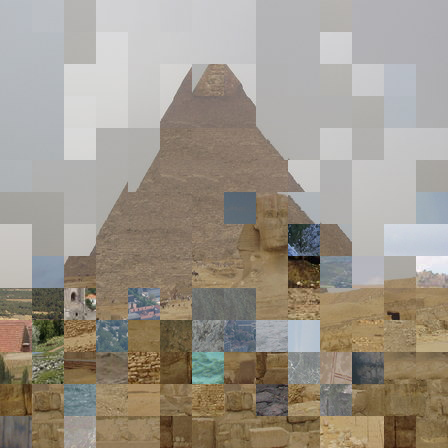} &
        \includegraphics[width=0.155\linewidth]{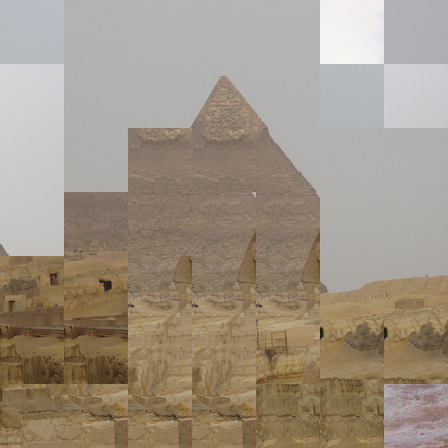} &
        \includegraphics[width=0.155\linewidth]{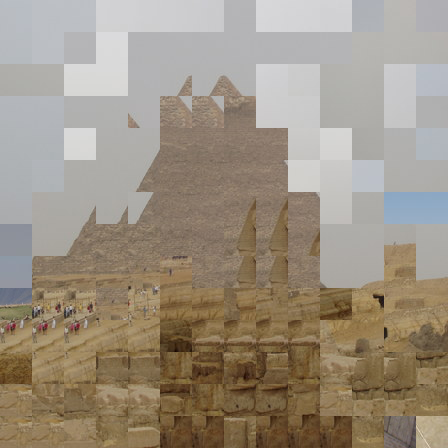} &
        \small{1-NN} \\
         &
        \includegraphics[width=0.155\linewidth]{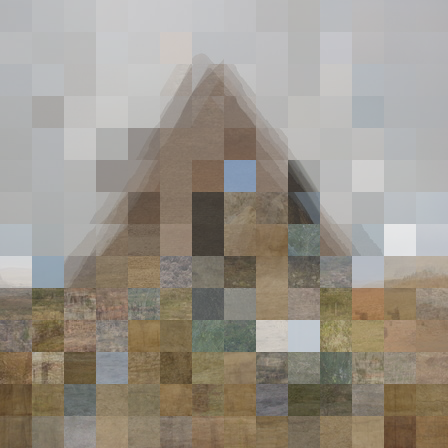} &
        \includegraphics[width=0.155\linewidth]{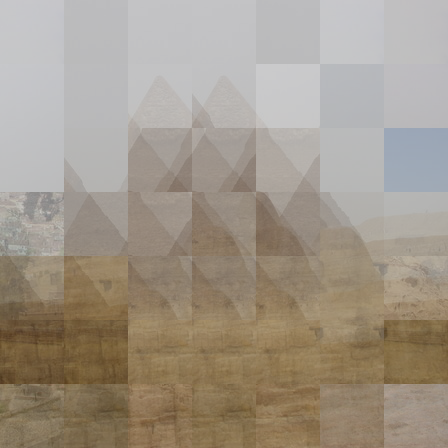} &
        \includegraphics[width=0.155\linewidth]{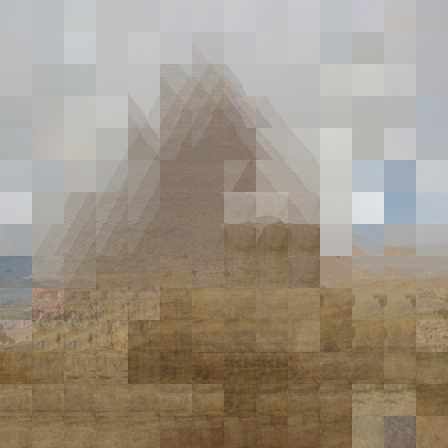} &
        \small{5-NN} \\

    \end{tabular}
\end{center}
\caption{Correspondence visualization of images (best viewed electronically).}
\label{fig:reconstruction}
\end{figure*}

\subsection{Feature visualization}
To further understand the features of different layers and scales, we produce visualizations of GoogLeNet features based on the Holidays dataset.

\textbf{Correspondence visualization}.
We construct a visualization to observe the correspondence behavior following~\cite{long2014convnets}.
% How it is done
%   1. Each local features represented by a patch
%      - n_l x n_l local features, having n/n_l x n/n_l
%   2. Find nearest neighbor patches in the database
%   3. Compute average
%  * Receptive fields are much larger, the patch can be seen as centers (+explanation)
To produce the visualization, we first represent each convolutional feature regarding a layer in the database by a square image patch which is obtained from the center of the image region that affects the local feature.
Specifically, for an $n \times n$ image with a layer output size $n^l \times n^l$, each local feature will be represented by a square image patch of size $\frac{n}{n^l} \times \frac{n}{n^l}$.
For each convolutional feature, the original image patch will be replaced by the average of its $k$ nearest neighbors from all patches extracted in the database.
If the local distinction has been abstracted by high level abstraction, locally different image patches will have similar neighbors as these patches may be semantically close; otherwise the neighbors can be also different since the local distinction is preserved.
Note that although the actual image region that affects the local features is much larger than the displayed patch itself due to stacked convolutions, the center patch still preserves localized correspondence~\cite{long2014convnets}.

% Why do you do this again? what is the difference between previous work?
%   1. Different architecture: GoogLeNet vs AlexNet
%   2. Investigating multiple scales
%   3. Their main thing is correspondence, we want to see how about instance (?)
%      - currently it seems it gives more semantically similar for high layer, though not local correspodence
% Why use this correspondence visualization?
%   1. Previous work use local information, robust to changes with little data (unlike classification)
%        - we believe this is important for instance retrieval
%   2. Also good for retrieval
The intermediate convolutional layers of the shallow AlexNet~\cite{krizhevsky2012imagenet} preserve correspondence between different instance objects as well as traditional SIFT descriptor~\cite{long2014convnets}.
However, as CNNs become deeper, it is unclear how the intermediate to high level convolutional layers would perform in capturing correspondence information.
% TODO mention explicitly about GoogLeNet?
In addition, we observe the behavior difference between scales of the feature from the visualization.
In particular, we would like to understand why the higher layers at finer scale obtain better performance than at lower scale.
\cite{long2014convnets} focuses on part correspondence across different object instances, which is in contrast to our goal of finding correspondence between objects.
However, we believe part correspondence is an important step for achieving instance correspondence, and this visualization is also useful in understanding the CNN features in instance correspondence.

The visualization is presented in Figure~\ref{fig:reconstruction}.
%
% Observations: What have we learned?
%   0a. Different strides: 7x7 vs 14x14
%   0b. Same strides for inception_5b-s2 vs inception_4a-s1
%   1. High layers have less correspondence: repetitive replacement
%       - Average Pooling after Inception 5b
%   2. Lower layers is distinctive
%       - less textureless region replaced by semantically irrelevant, like trees or grass
%       - still useful in retrieval, because it provides more original appearence distinctive information
%
The size of the convolutional feature map in \name{Inception 5b} scale 1 is $7 \times 7$, which is much smaller than $14 \times 14$ in \name{Inception 4a}'s .
Therefore, each patch of \name{Inception 5b} in the visualization is much larger than \name{Inception 4a}.
From the visualization, it is clear that features from \name{Inception 5b} do not correspond well compared to those from \name{Inception 4a}.
In \name{Inception 5b}, we can see many repetitive patterns for both 1-NN and 5-NN cases, which means that local features spatially close to each other are highly similar while the local appearance disparity between them is blurred by convolution operations.
One possible reason is that GoogLeNet is trained with average pooling just before softmax, which encourages the features of the last convolutional layer to be similar.
Comparing \name{Inception 5b} (scale 2) to \name{Inception 4a}, which have the same feature map sizes, \name{Inception 5b} retrieves more semantically relevant rather than locally distinct patches.
When applied to finer scale (scale 2), \name{Inception 5b} contains more local appearance details than the original scale, thus producing more diverse patches and roughly preserving the original appearance of the objects.
The visualization of \name{Inception 4a} contains more semantically irrelevant patches, especially in textureless regions, like retrieving grass or sea patches in the pyramid.
However, there are less repetitive patterns in the visualization, and the edges in the images are better preserved.
This shows that, as an intermediate convolutional layer, \name{Inception 4a} is more powerful at preserving correspondence of objects and capturing local appearance distinctions.

\begin{figure}[ht]
\begin{center}

\begin{subfigure}[ht]{\linewidth}
\begin{center}
\vspace{6px}
\includegraphics[width=0.08\linewidth]{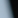}
\includegraphics[width=0.08\linewidth]{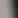}
\includegraphics[width=0.08\linewidth]{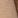}
\includegraphics[width=0.08\linewidth]{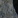}
\includegraphics[width=0.08\linewidth]{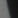}
\includegraphics[width=0.08\linewidth]{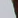}
\includegraphics[width=0.08\linewidth]{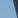}
\includegraphics[width=0.08\linewidth]{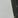}
\includegraphics[width=0.08\linewidth]{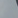}
\includegraphics[width=0.08\linewidth]{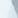}
\\
\includegraphics[width=0.08\linewidth]{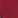}
\includegraphics[width=0.08\linewidth]{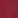}
\includegraphics[width=0.08\linewidth]{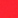}
\includegraphics[width=0.08\linewidth]{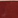}
\includegraphics[width=0.08\linewidth]{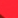}
\includegraphics[width=0.08\linewidth]{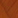}
\includegraphics[width=0.08\linewidth]{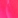}
\includegraphics[width=0.08\linewidth]{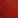}
\includegraphics[width=0.08\linewidth]{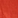}
\includegraphics[width=0.08\linewidth]{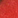}
\\
\includegraphics[width=0.08\linewidth]{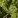}
\includegraphics[width=0.08\linewidth]{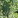}
\includegraphics[width=0.08\linewidth]{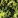}
\includegraphics[width=0.08\linewidth]{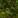}
\includegraphics[width=0.08\linewidth]{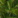}
\includegraphics[width=0.08\linewidth]{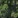}
\includegraphics[width=0.08\linewidth]{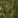}
\includegraphics[width=0.08\linewidth]{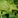}
\includegraphics[width=0.08\linewidth]{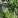}
\includegraphics[width=0.08\linewidth]{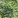}
\\
\includegraphics[width=0.08\linewidth]{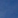}
\includegraphics[width=0.08\linewidth]{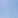}
\includegraphics[width=0.08\linewidth]{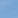}
\includegraphics[width=0.08\linewidth]{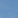}
\includegraphics[width=0.08\linewidth]{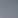}
\includegraphics[width=0.08\linewidth]{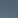}
\includegraphics[width=0.08\linewidth]{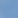}
\includegraphics[width=0.08\linewidth]{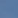}
\includegraphics[width=0.08\linewidth]{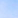}
\includegraphics[width=0.08\linewidth]{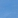}
\\
\includegraphics[width=0.08\linewidth]{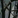}
\includegraphics[width=0.08\linewidth]{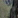}
\includegraphics[width=0.08\linewidth]{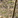}
\includegraphics[width=0.08\linewidth]{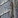}
\includegraphics[width=0.08\linewidth]{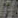}
\includegraphics[width=0.08\linewidth]{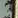}
\includegraphics[width=0.08\linewidth]{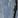}
\includegraphics[width=0.08\linewidth]{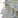}
\includegraphics[width=0.08\linewidth]{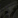}
\includegraphics[width=0.08\linewidth]{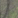}
\\
\caption{\name{Inception 3a} (scale 1)}
\end{center}
\end{subfigure}
\begin{subfigure}[ht]{\linewidth}
\begin{center}
\vspace{6px}
\includegraphics[width=0.08\linewidth]{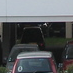}
\includegraphics[width=0.08\linewidth]{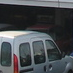}
\includegraphics[width=0.08\linewidth]{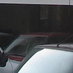}
\includegraphics[width=0.08\linewidth]{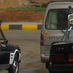}
\includegraphics[width=0.08\linewidth]{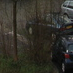}
\includegraphics[width=0.08\linewidth]{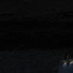}
\includegraphics[width=0.08\linewidth]{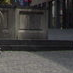}
\includegraphics[width=0.08\linewidth]{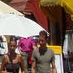}
\includegraphics[width=0.08\linewidth]{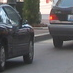}
\includegraphics[width=0.08\linewidth]{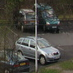}
\\
\includegraphics[width=0.08\linewidth]{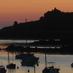}
\includegraphics[width=0.08\linewidth]{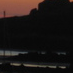}
\includegraphics[width=0.08\linewidth]{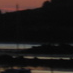}
\includegraphics[width=0.08\linewidth]{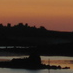}
\includegraphics[width=0.08\linewidth]{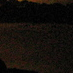}
\includegraphics[width=0.08\linewidth]{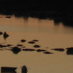}
\includegraphics[width=0.08\linewidth]{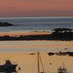}
\includegraphics[width=0.08\linewidth]{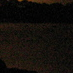}
\includegraphics[width=0.08\linewidth]{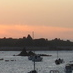}
\includegraphics[width=0.08\linewidth]{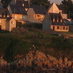}
\\
\includegraphics[width=0.08\linewidth]{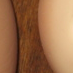}
\includegraphics[width=0.08\linewidth]{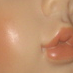}
\includegraphics[width=0.08\linewidth]{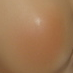}
\includegraphics[width=0.08\linewidth]{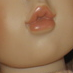}
\includegraphics[width=0.08\linewidth]{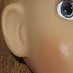}
\includegraphics[width=0.08\linewidth]{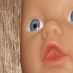}
\includegraphics[width=0.08\linewidth]{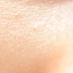}
\includegraphics[width=0.08\linewidth]{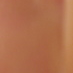}
\includegraphics[width=0.08\linewidth]{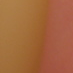}
\includegraphics[width=0.08\linewidth]{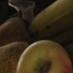}
\\
\includegraphics[width=0.08\linewidth]{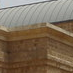}
\includegraphics[width=0.08\linewidth]{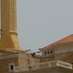}
\includegraphics[width=0.08\linewidth]{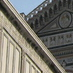}
\includegraphics[width=0.08\linewidth]{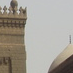}
\includegraphics[width=0.08\linewidth]{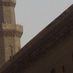}
\includegraphics[width=0.08\linewidth]{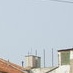}
\includegraphics[width=0.08\linewidth]{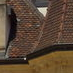}
\includegraphics[width=0.08\linewidth]{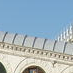}
\includegraphics[width=0.08\linewidth]{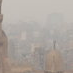}
\includegraphics[width=0.08\linewidth]{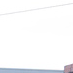}
\\
\includegraphics[width=0.08\linewidth]{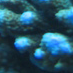}
\includegraphics[width=0.08\linewidth]{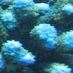}
\includegraphics[width=0.08\linewidth]{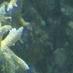}
\includegraphics[width=0.08\linewidth]{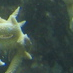}
\includegraphics[width=0.08\linewidth]{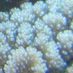}
\includegraphics[width=0.08\linewidth]{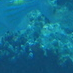}
\includegraphics[width=0.08\linewidth]{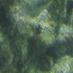}
\includegraphics[width=0.08\linewidth]{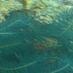}
\includegraphics[width=0.08\linewidth]{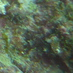}
\includegraphics[width=0.08\linewidth]{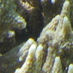}
\\
\caption{\name{Inception 5b} (scale 1)}
\end{center}
\end{subfigure}
\begin{subfigure}[ht]{\linewidth}
\begin{center}
\vspace{6px}
\includegraphics[width=0.08\linewidth]{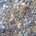}
\includegraphics[width=0.08\linewidth]{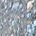}
\includegraphics[width=0.08\linewidth]{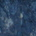}
\includegraphics[width=0.08\linewidth]{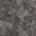}
\includegraphics[width=0.08\linewidth]{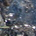}
\includegraphics[width=0.08\linewidth]{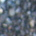}
\includegraphics[width=0.08\linewidth]{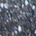}
\includegraphics[width=0.08\linewidth]{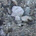}
\includegraphics[width=0.08\linewidth]{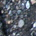}
\includegraphics[width=0.08\linewidth]{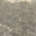}
\\
\includegraphics[width=0.08\linewidth]{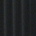}
\includegraphics[width=0.08\linewidth]{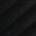}
\includegraphics[width=0.08\linewidth]{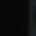}
\includegraphics[width=0.08\linewidth]{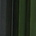}
\includegraphics[width=0.08\linewidth]{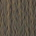}
\includegraphics[width=0.08\linewidth]{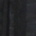}
\includegraphics[width=0.08\linewidth]{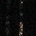}
\includegraphics[width=0.08\linewidth]{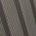}
\includegraphics[width=0.08\linewidth]{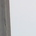}
\includegraphics[width=0.08\linewidth]{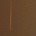}
\\
\includegraphics[width=0.08\linewidth]{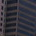}
\includegraphics[width=0.08\linewidth]{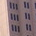}
\includegraphics[width=0.08\linewidth]{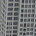}
\includegraphics[width=0.08\linewidth]{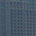}
\includegraphics[width=0.08\linewidth]{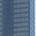}
\includegraphics[width=0.08\linewidth]{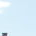}
\includegraphics[width=0.08\linewidth]{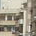}
\includegraphics[width=0.08\linewidth]{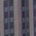}
\includegraphics[width=0.08\linewidth]{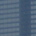}
\includegraphics[width=0.08\linewidth]{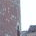}
\\
\includegraphics[width=0.08\linewidth]{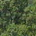}
\includegraphics[width=0.08\linewidth]{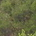}
\includegraphics[width=0.08\linewidth]{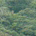}
\includegraphics[width=0.08\linewidth]{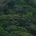}
\includegraphics[width=0.08\linewidth]{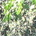}
\includegraphics[width=0.08\linewidth]{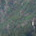}
\includegraphics[width=0.08\linewidth]{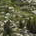}
\includegraphics[width=0.08\linewidth]{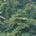}
\includegraphics[width=0.08\linewidth]{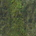}
\includegraphics[width=0.08\linewidth]{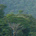}
\\
\includegraphics[width=0.08\linewidth]{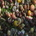}
\includegraphics[width=0.08\linewidth]{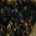}
\includegraphics[width=0.08\linewidth]{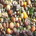}
\includegraphics[width=0.08\linewidth]{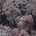}
\includegraphics[width=0.08\linewidth]{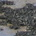}
\includegraphics[width=0.08\linewidth]{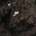}
\includegraphics[width=0.08\linewidth]{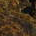}
\includegraphics[width=0.08\linewidth]{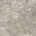}
\includegraphics[width=0.08\linewidth]{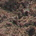}
\includegraphics[width=0.08\linewidth]{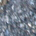}
\\
\caption{\name{Inception 5b} (scale 2)}
\end{center}
\end{subfigure}

\end{center}
\caption{Visualization of local convolutional features on different layers and scales.
Each row represents a cluster of local convolutional features by displaying the corresponding patches.
The leftmost column shows the sampled reference patches, and other patches are sorted according to their L2 distance with the reference patches.
}
\label{fig:patches}
\end{figure}

%%%%%%%%%%%%%%%%%%%%%%%%%%%%%%%%%%%%%%%%%%%%%%%%%%%
\textbf{Patch clusters}.
To better observe the clustering of local CNN features, we sample patches in the dataset and show their nearest neighbors on different layers.
Each convolutional feature is represented as a patch in the same way as in the correspondence visualization.
Figure~\ref{fig:patches} shows the patch clustering visualization of GoogLeNet layers \name{Inception 3a}, \name{Inception 5b} and \name{Inception 5b} (scale 2).
The patch clusters in the lower layer \name{Inception 3a} are quite similar to SIFT-like low level features, where strong edges, corners and texture are discovered and encoded.
For higher layers, such as \name{Inception 5b}, we can see more generalization of parts with semantic meaning, such as different views of a car or scene, which reflects the tendency of higher layers to capture category-level invariances.
However, for the same layer \name{Inception 5b} applied to the finer scale, the features focus on smaller parts of the images, thus capturing more local appearance.
This confirms that the features behave quite differently when applied to images of different resolutions.
Although the higher layers are supposed to encode high level categorical features, more instance-level details are also preserved when they are applied to finer scales, so they are more useful for image retrieval.
%

% generated by code/grab_patches_clusters.py

%
%Similar to SIFT, the convolutional features can be regarded as a local feature, which can be matched.
%Figure ? shows matching patches.
%
\begin{table}[ht]
\begin{center}
{\small
\begin{tabular}{ c|c c c}
\hline
 Method & Holidays & Oxford & Paris\\
\hline\hline
%CNN-ss \cite{razavian2014cnn}                             & 76.9  & ? \\
\multicolumn{4}{c}{SIFT-based method} \\
\hline
BoW 200k-D~\cite{DBLP:journals/pami/JegouPDSPS12-vlad} & 54.0 & 36.4 & 46.0 \\
Improved Fisher~\cite{DBLP:conf/cvpr/PerronninLSP10} & 62.6 & 41.4 & - \\
LCS+RN~\cite{DBLP:conf/mm/DelhumeauGJP13} & 65.8 & 51.7 & - \\
VLADintra+ RootSIFT~\cite{arandjelovic2013all}                       & 65.3  & 55.8 & -\\
CVLAD~\cite{DBLP:conf/bmvc/ZhaoGJ13} & 82.7 & 51.4 & - \\
\hline
\multicolumn{4}{c}{CNN-based method} \\
\hline
CNNaug-ss \cite{razavian2014cnn}                          & 84.3  & 68.0 & 79.5\\
Multi-resolution & \multirow{2}{*}{\bf 89.7} & \multirow{2}{*}{\bf 84.4} & \multirow{2}{*}{\bf 85.3} \\
Spatial Search \cite{razavian2015visual}            &  & & \\
Neural codes~\cite{babenko2014neural}         & 79.3  & 54.5 & -\\
MOP-CNN~\cite{gong2014multi} & 80.2  & - & -\\
\hline
\hline
Ours (OxfordNet) & 83.8 & 64.9 & 69.4\\
Ours (GoogLeNet) & 84.0 & 58.1 & 68.8\\
\hline
\end{tabular}
}
\end{center}
\caption{Comparison with other methods on image retrieval dataset.}
\label{table:compare}
\end{table}

%\vspace{-5mm}
\subsection{Comparison to state-of-the-art}

Since our method only uses simple CNN features and VLAD encoding, we only compare to other recent CNN based approaches and classic SIFT-based representations with BoW and VLAD encoding.

\textbf{Uncompressed representation}.
We first compare our approach using uncompressed VLAD representation with other state-of-the-art approaches in Table~\ref{table:compare}.
In Figure~\ref{fig:layers}, the best performing layers on Holidays, Oxford and Paris datasets are \name{Inception 3a} on original scale (scale 1), \name{Inception 5b} and \name{Inception 4e} on finer scale (scale 2) on GoogLeNet respectively, and \name{conv4\_2}, \name{conv5\_1} and \name{conv5\_2} for Holidays, Oxford and Paris dataset on OxfordNet respectively.
The VLAD descriptors from the two scales on the best performing layer are concatenated as our final multi-scale descriptors.
OxfordNet, which has much larger convolutional feature maps, performs slightly better than GoogLeNet for image retrieval.
Although we do not focus on producing state-of-the-art results on image retrieval but more on investigating the behavior of convolutional features from different layers and the effect of multiple scales, our system gives competitive results compared to state-of-the-art methods.
Specifically, our approach significantly outperforms all the classic SIFT-based approaches with BoW and VLAD encoding, which verifies the representative power of the convolutional features compared to traditional SIFT descriptors.
Although better results are reported by other SIFT-based approaches using large vocabularies, spatial verification and query expansion, etc., our framework is not limited to the current setting, and can be readily adapted to other encoding schemes (\emph{i.e.}, BoW and FV), and re-ranking techniques (\emph{i.e.}, query expansion).
In addition, compared to recent CNN-based approaches, our method still produces better or comparable results.
In particular, our approach outperforms its rivals that either use time-consuming multi-scale sliding windows to extract features~\cite{gong2014multi} or retrain the entire network using extra data~\cite{babenko2014neural}.
It should be noted that including spatial information greatly boosts the performance of CNN-based approaches such as spatial search~\cite{razavian2014cnn, razavian2015visual}.
Although \cite{razavian2014cnn} and \cite{razavian2015visual} produce better results than our method, we believe that our approach of extracting and encoding convolutional features using lower layers and our investigation of how scales affect convolutional features provide a better understanding of why spatial search on multi-scale features from the last layer performs well.
Spatial information can be also included in our framework with few modifications, which will be studied in future work.
It would also be interesting to combine multiple layers from the best scales in spatial search to fully utilize the power of deep networks.
\begin{table}
\begin{center}
{\small
\begin{tabular}{ c|c|c c c c}
\hline
Method & dim & Holidays & Oxford & Paris \\
\hline\hline
%CNN-ss \cite{razavian2014cnn}     &                        & 76.9  & ?  & ?\\
VLADintra+SIFT~\cite{arandjelovic2013all} & 128 & 62.5 & 44.8 & -\\
%gVLAD~\cite{DBLP:journals/corr/WangDBJP14} & 128 &  77.9 & 60.0 \\
FV+T-embedding~\cite{DBLP:conf/cvpr/JegouZ14} & 128 & 61.7 & 43.3 & - \\
Neural codes \cite{babenko2014neural} & 128 & 78.9  & 55.7 & - \\
MOP-CNN \cite{gong2014multi} & 512 & 78.4  & - & - \\
Spatial Pooling \cite{razavian2015visual} & 256 & 74.2 & 53.3 & \textbf{67.0} \\
\hline
\hline
Ours (OxfordNet) & 128 & 81.6 & \textbf{59.3} & 59.0 \\
Ours (GoogLeNet) & 128 & \textbf{83.6} & 55.8 & 58.3 \\
\hline
\end{tabular}
}
\end{center}
\caption{Comparison of low dimensional descriptors.}
\label{table:compare_small}
\end{table}

\textbf{Low-dimensional representation.}
To trade-off between retrieval accuracy and storage space, most approaches compress the original feature vector to a low-dimensional representation.
Therefore, we conduct additional experiments using compressed VLAD descriptors and compare the results with those of other approaches using low-dimensional representations.
We use PCA to reduce the dimensionality to 128 and apply whitening to further remove noise.

As shown in Table~\ref{table:compare_small}, our method obtains state-of-the-art results on two out of three datasets with minimal performance loss.
Our method outperforms all SIFT-based approaches by a large margin, which again demonstrates the power of CNNs.
Moreover, we obtain better results than \cite{babenko2014neural}, even though \cite{babenko2014neural} fine-tunes the pre-trained CNNs using a large amount of additional data.
Although adopting similar VLAD encoding scheme, our method still outperforms MOP-CNN~\cite{gong2014multi} which uses a larger 512-D representation, which further verifies that our approach of extracting convolutional features from intermediate layers is more suitable for instance-level image retrieval.
The performance of \cite{razavian2015visual} with low-dimensional descriptors drops notably compared to our 128-D representation, showing that elimination of spatial search greatly reduces the power of CNN representation.
It is also important to use more sophisticated encoding methods to capture the local information of convolutional features instead of simple max-pooling as in~\cite{razavian2015visual}.
In contrast, our low-dimensional representation is robust and retains good discriminative power.

\section{Conclusion}
In this work, we systematically experiment with features from different layers of convolutional networks and different scales of input images for instance-level image retrieval, and provide insights into performance through various visualizations.
%We proposed a novel way to encode convolutional responses with VLAD, and achieve state-of-the-art retrieval results using low dimensional representations on two of the instance image retrieval datasets.
With VLAD encoding on convolutional response, we achieve state-of-the-art retrieval results using low dimensional representations on two of the instance image retrieval datasets.

\clearpage
{\small
\bibliographystyle{ieee}
\bibliography{egbib}

\begin{thebibliography}{10}\itemsep=-1pt

\bibitem{DBLP:conf/cvpr/ArandjelovicZ12}
R.~Arandjelovi\'c and A.~Zisserman.
\newblock Three things everyone should know to improve object retrieval.
\newblock In {\em CVPR}, pages 2911--2918, 2012.

\bibitem{arandjelovic2013all}
R.~Arandjelovic and A.~Zisserman.
\newblock All about {VLAD}.
\newblock In {\em {CVPR}}, pages 1578--1585, 2013.

\bibitem{babenko2014neural}
A.~Babenko, A.~Slesarev, A.~Chigorin, and V.~Lempitsky.
\newblock Neural codes for image retrieval.
\newblock In {\em {ECCV}}, pages 584--599. 2014.

\bibitem{Crowley14a}
E.~J. Crowley and A.~Zisserman.
\newblock In search of art.
\newblock In {\em Workshop on Computer Vision for Art Analysis, ECCV}, 2014.

\bibitem{DBLP:conf/mm/DelhumeauGJP13}
J.~Delhumeau, P.~H. Gosselin, H.~J{\'{e}}gou, and P.~P{\'{e}}rez.
\newblock Revisiting the {VLAD} image representation.
\newblock In {\em {ACM} Multimedia}, pages 653--656, 2013.

\bibitem{DBLP:conf/icml/DonahueJVHZTD14}
J.~Donahue, Y.~Jia, O.~Vinyals, J.~Hoffman, N.~Zhang, E.~Tzeng, and T.~Darrell.
\newblock {DeCAF}: {A} deep convolutional activation feature for generic visual
  recognition.
\newblock In {\em {ICML}}, pages 647--655, 2014.

\bibitem{DBLP:conf/cvpr/DouzeRS11}
M.~Douze, A.~Ramisa, and C.~Schmid.
\newblock Combining attributes and fisher vectors for efficient image
  retrieval.
\newblock In {\em {CVPR}}, pages 745--752, 2011.

\bibitem{gong2014multi}
Y.~Gong, L.~Wang, R.~Guo, and S.~Lazebnik.
\newblock Multi-scale orderless pooling of deep convolutional activation
  features.
\newblock In {\em {ECCV}}, pages 392--407. 2014.

\bibitem{DBLP:conf/eccv/JegouC12}
H.~J{\'e}gou and O.~Chum.
\newblock Negative evidences and co-occurences in image retrieval: The benefit
  of {PCA} and whitening.
\newblock In {\em ECCV}, pages 774--787, 2012.

\bibitem{Jegou2008hamming}
H.~J\'egou, M.~Douze, and C.~Schmid.
\newblock Hamming embedding and weak geometric consistency for large scale
  image search.
\newblock In {\em ECCV}, pages 304--317, 2008.

\bibitem{DBLP:journals/pami/JegouPDSPS12-vlad}
H.~J{\'e}gou, F.~Perronnin, M.~Douze, J.~S{\'a}nchez, P.~P{\'e}rez, and
  C.~Schmid.
\newblock Aggregating local image descriptors into compact codes.
\newblock {\em IEEE Trans. Pattern Anal. Mach. Intell.}, 34(9):1704--1716,
  2012.

\bibitem{DBLP:conf/cvpr/JegouZ14}
H.~J{\'{e}}gou and A.~Zisserman.
\newblock Triangulation embedding and democratic aggregation for image search.
\newblock In {\em {CVPR}}, pages 3310--3317, 2014.

\bibitem{jia2014caffe}
Y.~Jia, E.~Shelhamer, J.~Donahue, S.~Karayev, J.~Long, R.~Girshick,
  S.~Guadarrama, and T.~Darrell.
\newblock Caffe: Convolutional architecture for fast feature embedding.
\newblock {\em arXiv preprint arXiv:1408.5093}, 2014.

\bibitem{DBLP:conf/iccv/KovashkaG13a}
A.~Kovashka and K.~Grauman.
\newblock Attribute adaptation for personalized image search.
\newblock In {\em {ICCV}}, pages 3432--3439, 2013.

\bibitem{DBLP:conf/iccv/KovashkaG13}
A.~Kovashka and K.~Grauman.
\newblock Attribute pivots for guiding relevance feedback in image search.
\newblock In {\em {ICCV}}, pages 297--304, 2013.

\bibitem{DBLP:conf/cvpr/KovashkaPG12}
A.~Kovashka, D.~Parikh, and K.~Grauman.
\newblock Whittlesearch: Image search with relative attribute feedback.
\newblock In {\em {CVPR}}, pages 2973--2980, 2012.

\bibitem{krizhevsky2012imagenet}
A.~Krizhevsky, I.~Sutskever, and G.~E. Hinton.
\newblock {ImageNet} classification with deep convolutional neural networks.
\newblock In {\em {NIPS}}, pages 1097--1105, 2012.

\bibitem{long2014convnets}
J.~Long, N.~Zhang, and T.~Darrell.
\newblock Do convnets learn correspondence?
\newblock In {\em {NIPS}}, pages 1601--1609, 2014.

\bibitem{DBLP:conf/cvpr/OquabBLS14}
M.~Oquab, L.~Bottou, I.~Laptev, and J.~Sivic.
\newblock Learning and transferring mid-level image representations using
  convolutional neural networks.
\newblock In {\em {CVPR}}, pages 1717--1724, 2014.

\bibitem{DBLP:conf/iccv/ParikhG13}
D.~Parikh and K.~Grauman.
\newblock Implied feedback: Learning nuances of user behavior in image search.
\newblock In {\em {ICCV}}, pages 745--752, 2013.

\bibitem{DBLP:conf/cvpr/PerronninLSP10}
F.~Perronnin, Y.~Liu, J.~S{\'{a}}nchez, and H.~Poirier.
\newblock Large-scale image retrieval with compressed fisher vectors.
\newblock In {\em {CVPR}}, pages 3384--3391, 2010.

\bibitem{DBLP:conf/cvpr/PhilbinCISZ07}
J.~Philbin, O.~Chum, M.~Isard, J.~Sivic, and A.~Zisserman.
\newblock Object retrieval with large vocabularies and fast spatial matching.
\newblock In {\em CVPR}, pages 1--8, 2007.

\bibitem{DBLP:conf/cvpr/PhilbinCISZ08}
J.~Philbin, O.~Chum, M.~Isard, J.~Sivic, and A.~Zisserman.
\newblock Lost in quantization: Improving particular object retrieval in large
  scale image databases.
\newblock In {\em {CVPR}}, 2008.

\bibitem{DBLP:conf/cvpr/RastegariDPF13}
M.~Rastegari, A.~Diba, D.~Parikh, and A.~Farhadi.
\newblock Multi-attribute queries: To merge or not to merge?
\newblock In {\em {CVPR}}, pages 3310--3317, 2013.

\bibitem{razavian2014cnn}
A.~S. Razavian, H.~Azizpour, J.~Sullivan, and S.~Carlsson.
\newblock Cnn features off-the-shelf: an astounding baseline for recognition.
\newblock In {\em {CVPR} Workshops}, pages 512--519, 2014.

\bibitem{razavian2015visual}
A.~S. Razavian, J.~Sullivan, A.~Maki, and S.~Carlsson.
\newblock Visual instance retrieval with deep convolutional networks.
\newblock {\em CoRR}, abs/1412.6574, 2014.

\bibitem{DBLP:journals/corr/SermanetEZMFL13}
P.~Sermanet, D.~Eigen, X.~Zhang, M.~Mathieu, R.~Fergus, and Y.~LeCun.
\newblock Overfeat: Integrated recognition, localization and detection using
  convolutional networks.
\newblock {\em CoRR}, abs/1312.6229, 2013.

\bibitem{DBLP:conf/cvpr/SiddiquieFD11}
B.~Siddiquie, R.~S. Feris, and L.~S. Davis.
\newblock Image ranking and retrieval based on multi-attribute queries.
\newblock In {\em {CVPR}}, pages 801--808, 2011.

\bibitem{Simonyan14c}
K.~Simonyan and A.~Zisserman.
\newblock Very deep convolutional networks for large-scale image recognition.
\newblock {\em CoRR}, abs/1409.1556, 2014.

\bibitem{DBLP:conf/iccv/SivicZ03}
J.~Sivic and A.~Zisserman.
\newblock Video {Google}: A text retrieval approach to object matching in
  videos.
\newblock In {\em ICCV}, pages 1470--1477, 2003.

\bibitem{szegedy14going}
C.~Szegedy, W.~Liu, Y.~Jia, P.~Sermanet, S.~Reed, D.~Anguelov, D.~Erhan,
  V.~Vanhoucke, and A.~Rabinovich.
\newblock Going deeper with convolutions.
\newblock {\em CoRR}, abs/1409.4842, 2014.

\bibitem{DBLP:conf/mm/WanWHWZZL14}
J.~Wan, D.~Wang, S.~C.~H. Hoi, P.~Wu, J.~Zhu, Y.~Zhang, and J.~Li.
\newblock Deep learning for content-based image retrieval: {A} comprehensive
  study.
\newblock In {\em {ACM Multimedia}}, pages 157--166, 2014.

\bibitem{DBLP:journals/corr/XuYH14}
Z.~Xu, Y.~Yang, and A.~G. Hauptmann.
\newblock A discriminative {CNN} video representation for event detection.
\newblock {\em arXiv preprint arXiv:1411.4006}, 2014.

\bibitem{DBLP:conf/eccv/ZeilerF14}
M.~D. Zeiler and R.~Fergus.
\newblock Visualizing and understanding convolutional networks.
\newblock In {\em {ECCV}}, pages 818--833, 2014.

\bibitem{DBLP:conf/bmvc/ZhaoGJ13}
W.~Zhao, G.~Gravier, and H.~J{\'{e}}gou.
\newblock Oriented pooling for dense and non-dense rotation-invariant features.
\newblock In {\em {BMVC}}, 2013.

\end{thebibliography}
}

\end{document}